\documentclass[10pt]{article}
\usepackage{amsfonts}
\usepackage{epsfig}

\usepackage{color}

\begin{document}

\title{The settlement of Madagascar: what dialects and languages
can tell}
\author{Maurizio Serva}
\maketitle
\centerline{\it Dipartimento di Matematica, Universit\`{a} dell'Aquila, 
I-67010 L'Aquila, Italy}

\begin{abstract}

The dialects of Madagascar belong to the Greater Barito East group of the
Austronesian family and it is widely accepted that the Island
was colonized by Indonesian sailors after a maritime trek which probably 
took place around 650 CE.
The language most closely related to Malagasy dialects is Maanyan but 
also Malay is strongly related especially for what concerns navigation terms.
Since the Maanyan Dayaks live along the Barito river in Kalimantan
(Borneo) and they do not possess the necessary skill for long maritime
navigation, probably they were brought as subordinates by Malay sailors.

In a recent paper we compared 23 different Malagasy dialects in order to 
determine the time and the landing area of the first colonization.
In this research we use new data and new methods to confirm that
the landing took place on the south-east coast of the Island.
Furthermore, we are able to state here that it is unlikely that
there were multiple settlements and, therefore, colonization consisted in
a single founding event.

To reach our goal we find out the internal kinship relations among all
the 23 Malagasy dialects and we also find out the different kinship 
degrees of the 23 dialects versus Malay and Maanyan. 
The method used is an automated version of the lexicostatistic approach.
The data concerning Madagascar were collected by the author 
at the beginning of 2010 and consist of Swadesh lists 
of 200 items for 23 dialects covering all areas of the Island. 
The lists for Maanyan and Malay were obtained from published datasets 
integrated by author's interviews.

\end{abstract}

\bigskip

\begin{flushleft}
\end{flushleft}

\section*{Introduction}

Malagasy language (as well all its dialects) belongs to 
the Austronesian linguistic family. This was definitively established in 
\cite{Dahl} where it is also shown a particularly close relationship 
between Malagasy and Maanyan which is spoken by a Dayak community of Borneo.
A relevant contribution also comes from loanwords of other Indonesian languages
as Ngaju Dayak, Buginese, Javanese and Malay \cite{Ad1,Ad2}. 
In particular, Malay is very well represented in the domain of navigation terms. 
A very small amount of the vocabulary
can be associated with non-Austronesian languages (for example
Bantu languages for what concerns faunal names \cite{BW}).

The Indonesian colonizers reached Madagascar by a maritime trek 
at a time that we estimated in a recent paper \cite{SPVW}
to be around 650 CE, a date which is within the widely accepted
range of time \cite{Ad1,Ad2}.
In the same paper we found a strong indication that the landing area 
was in the south-east of the Island.
This was established assuming that the homeland is the area
exhibiting the maximum of current linguistic diversity. 
Diversity was measured by comparing lexical and geographical distances. 

In this paper we confirm the south-east location as the area of
landing (were the population dispersal took origin).
Furthermore, we find out that colonization consisted in
a single founding event. Therefore, it is unlikely that
there were multiple settlements and eventual subsequent
landings did not alter consistently the linguistic equilibrium.
Our study starts from the consideration
that Maanyan speakers, which live along the rivers 
of Kalimantan, do not have the necessary skills for long-distance 
maritime navigation. The most reasonable explanation \cite{Ad1,Ad2}
is that they were brought as subordinates by Malay sailors. 
For this reason we reexamine the internal kinship relations among all
the 23 Malagasy dialects but we also perform a comparison 
of all these variants with respect both Malay and Maanyan.
These new output concerning Malagasy dialects and their
relations with the two Indonesian languages
are examined with new methods which all confirm that
the landing took place on the south-east coast of the Island.

The vocabulary used for the present study
was collected by the author with the invaluable help of 
Joselin\`a Soafara N\'er\'e at the beginning of 2010. 
The dataset, which can be found in \cite{SP2}, 
consists of 200 words Swadesh lists \cite{Sw} for 23
dialects of Malagasy from all the areas of the Island.
The orthographical conventions are those of standard Malagasy.
Most of the informants were able to write the words directly 
using these conventions, while a few of them benefited from the help of 
one ore more fellow townsmen. 
For any dialect list two different speakers have been interviewed,
their complete list is provided in Appendix B
while the locations can be seen in Fig. 2.
Finally, the lists for Maanyan and Malay were 
obtained by published dataset \cite{Gre} integrated by author's interviews.

\section*{Method}

The method that we use \cite{SP1,PS1} is based on a lexical comparison 
of languages by means of an automated measure of distance 
between pairs of words with same meaning contained is their Swadesh lists.
The use of Swadesh lists \cite{Sw} in lexicostatistics is popular since 
half a century.  They are lists of words associated to the same $M$ meanings,
(the original Swadesh choice was $M=200$) which concern the basic activities 
of humans. 
Comparing the two lists corresponding to a pair of languages
it is possible to determine the percentage of shared {\it cognates}
which is a measure of their lexical distance.
A recent example of the use of Swadesh lists and cognates counting to 
construct language trees are the studies of Gray and Atkinson \cite{GA} 
and Gray and Jordan \cite{GJ}.

The idea of measuring relationships among languages
using vocabulary is much older than lexicostatistics
and it seems to have its roots in the work of 
the French explorer Dumont D'Urville.
He collected comparative word lists during his voyages aboard the Astrolabe 
from 1826 to 1829 and, in his work about 
the geographical division of the Pacific \cite{Urv},
he proposed a method to measure the degree of relation 
among languages.
He used a core vocabulary of 115 terms, then he assigned a distance 
from 0 to 1 to any pair of words with the 
same meaning and finally he was able to determine the degree of
relation between any pair of languages.

Our automated method (see Appendix A for details) works as follows:
for any language we write down a Swadesh list, then 
we compare words with same meaning belonging to
different languages only considering orthographic differences.
This approach is motivated by the analogy with genetics: 
the vocabulary has the role of DNA and the comparison is simply made by 
measuring the differences between the DNA of the two languages.
There are various advantages: 
the first is that, at variance with previous methods, 
it avoids subjectivity, the second is
that results can be replicated by other scholars assuming that the
database is the same, the third is that it is not requested a specific 
expertize in linguistic, and the last, but surely not the least, is
that it allows for a rapid comparison of a very large number of languages
(or dialects).

If a family of languages is considered, all the information is encoded 
in a matrix whose entries are the pairwise lexical distances, nevertheless,
this information is not manifest and it has to be extracted.  
The ubiquitous approach to this problem
is to transform the matrix information in a phylogenetic tree.

Nevertheless, in this transformation, part of the information may be lost
because transfer among languages is not exclusively vertical
(as in mtDNA transmission from mother to child)
but it also can be horizontal (borrowings and, in extreme cases, creolization).
Another approach is the geometric one \cite{Bl,SPVW} that results
from Structural Component Analysis (SCA) that we have recently proposed.
This approach encodes the matrix information into the positions of 
the languages in a $n$-dimensional space. 
For large $n$ one recovers all the matrix content, but a low dimensionality, 
typically $n$=2 or $n$=3, is sufficient to grasp all the relevant information.
The results in this paper mostly rely to a direct investigation of the 
entries of the matrix and to simple averages over them.

\section*{Malagasy dialects}

The number of Malagasy dialects we consider is $N$=23,
therefore, the output of our method, when applied only
to these variants is a 
matrix with $N(N-1)/2=253$ non-trivial entries
representing all the possible lexical distances among dialects.
This matrix is explicitly shown in Appendix A.

The information concerning the vertical transmission of vocabulary
from the proto-Malagasy to the contemporary dialects can be extracted by
a phylogenetic approach. There are various possible choices for the 
algorithm for the reconstruction of the family tree
(see \cite{SPVW} for a discussion of this point), 
we show in Fig. 1 the output of the Unweighted Pair 
Group Method Average (UPGMA).
In this figure the name of the dialect is followed
by the name of the town were it was collected.
The input data for the UPGMA tree are the pairwise 
separation times obtained from the lexical distances by means of a 
simple logarithmic rule (\cite{SP1, PS1}).
The absolute time-scale is calibrated by the results of the SCA analysis, 
which indicate a separation date 650 CE \cite{SPVW}. 
The phylogenetic tree in Fig. 1 interestingly shows a main 
partition of Malagasy dialects in two main branches (east-center-north 
and south-west) at variance with previous studies which gave a different
partitioning \cite{Ver} (indeed,
the results in \cite{Ver} coincide with ours if a correct
phylogeny is applied, see \cite{SPVW} for a discussion of this point.)   
Then, each of two branches splits, in turn, in two sub-branches
whose leaves are associated to different colors. 
In order to demonstrate the strict correspondence of this cladistic 
with the geography, we display a map of Madagascar (Fig. 2)
where the locations of the 23 dialects are indicated with the same 
colors of the leaves in Fig. 1.
We remark the relative isolation of the Antandroy variant (yellow).

Up to know, we only have shown the consistency of the approach 
which can be appreciated by comparison between Fig. 1 and Fig. 2.
We start our investigation by computing
the average distance of each of the dialects
from all the others (see Fig. 3).
Antandroy has the largest average distance, confirming that it is 
the overall most deviant variant (something which is also commonly 
pointed out by other Malagasy speakers). 
We further note that the smallest average distance is for Merina
(official language), Betsileo and Bara, which are all spoken 
on the highlands. 
The fact that the Merina has the smallest average distance is 
possibly partially explained by the fact that this variant is the 
official one. However, as we will show later by means of a comparison 
of Malagasy dialects with Malay and Maanyan, this cannot 
be the only explanation.
More interestingly we remark that the Antambohoaka and Antaimoro variants, 
which are spoken in Mananjary and Manakara also have a very small
average distance from the other dialects. Both this dialects
are spoken in the south-east coast of Madagascar in a 
relatively isolated position and, therefore, this is the first evidence for 
south-east as the homeland of the Malagasy language
and, likely, as the location of the first settlement.

The identification of the southeastern coast of Madagascar as the
landing area for the Indonesian colonizers is supported by 
geographical considerations. In fact, there is an Indian Ocean current 
which goes from Sumatra to Madagascar. 
When Mount Krakatoa erupted in 1883, pumice arrived on 
south-east coast where the Mananjary River opens into the sea.
Furthermore, during the Second World War, the same area saw the arrival of
pieces of wreckage from ships sailing between Java
and Sumatra that had been bombed by the Japanese air-force. 
Notice that the mouth of the Mananjary River is where
the town of Mananjary is presently located, and it is
also close to Manakara. 
The Indonesian ancestors of today Malagasy probably profited of this current,
which they possibly entered sailing throw the Sunda strait.

\section*{Dialects, Malay and Maanyan}

The classification of Malagasy (and its dialects) among the 
Greater Barito East languages of Borneo as well as the particularly close relationship 
with Maanyan is beyond doubt.
However, Malagasy also underwent influences from other Indonesian languages 
as Ngaju Dayak, Javanese, Buginese and, particularly, from Malay which
exhibits the most relevant relationship after Maanyan. 

If we consider the 23 dialects together with Malay and Maanyan, not only 
we have to compute the 253 internal distances, but also we have to determine
the 23x2=56 distances of any of the dialects from the two Indonesian languages.
These new distances are displayed in Fig. 4.

First of all we observe, as expected, that the largest of the distances 
from Maanyan is smaller then the smallest of the distances from Malay.
This simply reflects the fact that Malagasy is first of all an East 
Barito language.
Then we also observe that Malagasy dialects seem to have almost the 
same relative composition. In fact, all the points in Fig. 4 have 
almost the same {\it distance from Malay}/{\it distance from Maanyan} ratio.
This is a strong indication that the linguistic makeup
is substantially the same for all dialects
and, therefore, that they all originated by the same founding population 
of which they reflect the initial composition.
The conclusion is that the founding event
was likely a single one and subsequent immigration 
did not alter significantly the linguistic composition.

Indeed, looking more carefully, one can detect a little less Malay in the 
north since red circles have a larger ratio with 
respect all the others. This cannot a be a consequence of a larger 
African influence in the vocabulary due to the active trade with the continent 
and Comoros Islands. In this case both the Maanyan and Malay 
component of the vocabulary would be affected. 
Instead, this may be the effect of Malay trading which, according to Adelaar
\cite{Ad1, Ad2}, continued for several centuries after colonization. 
 
Noticeably, some dialects changed less with respect to the proto-language 
(Antananarivo, Fianarantsoa, Manajary, Manakara), in fact, their distances 
both from Maanyan and Malay are smaller then those of the other dialects. 
This is probably the most relevant phenomenon, and we 
underline that the variants which 
are less distant on average with respect to the other dialects 
(Fig. 3) are also less distant with respect to Malay and Maanyan (Fig. 4).
Therefore, the fact that Merina is closer to the other dialects cannot
be merely justified by the fact that it is the official variant.

We have checked whether the picture which emerges from Fig. 4
is confirmed by comparing with other related Indonesian languages.
The result is positive, and in particular the dialects of Manajary,
Manakara, Antananarivo and Fianarantsoa seem to be closer to 
most of the Indonesian languages which we compare them to.
Note that  Manajary and Manakara are both in the previously identified 
landing area on the south-east coast
while Antananarivo and Fianarantsoa are in the central highlands of the Island.
This suggests a scenario according to which there was a migration 
on the highlands of Madagascar (Betsileo and Imerina regions)
shortly after the landing on the south-east coast (Manakara, Manajary).

In conclusion, both average distances in Fig. 3 and distances from
related Indonesian language point to the south-east coast
as the area of the first settlement.
This is the same indication which comes from the fact that linguistic 
diversity is higher in that region (see \cite{SPVW}). 

Finally, we remark that the Antandroy variant (Ambovombe) 
is the most distant from Maanyan and 
among the most distant dialects from Malay, showing again to be the most 
deviant dialect. It is not clear whether its divergent evolution was due to 
internal factors or to specific language contacts which are still to be identified.

\section*{Outlook}

The main open problem concerning Malagasy is to determine the
composition of the population which settled the Island.
Adelaar writes : {\it Malay influence persisted for several
centuries after the migration. But, except for this Malay influence, most
influence on Malagasy from other Indonesian languages seems to be 
pre-migratory. (...) I also believe it possible that the
early migrants from south-east Asia came not exclusively from the 
south-east Barito area, in fact, that south-east Barito speakers may 
not even have constituted a majority among these migrants, but rather formed a
nuclear group which was later reinforced by south-east Asian migrants
with a possibly different linguistic and cultural background (and, of course,
by African migrants).
Whatever view one may hold on how the early Malagasy were
influenced by other Indonesians, it seems necessary that we at least
develop a more cosmopolitan view on the Indonesian origins of the
Malagasy. A south-east Barito origin is beyond dispute, but this is of
course only one aspect of what Malagasy dialects and cultures reflect
today. Later influences were manifold, and some of these influences,
African as well as Indonesian, were so strong that they have molded the
Malagasy language and culture in all its variety into something new,
something for the analysis of which a south-east Barito origin has become a
factor of little explanatory value.}
 
In order to clarify the problem raised by Adelaar, it is
necessary to understand the Malagasy relationships with other Indonesian
languages (and possibly African ones). 
The fact that the use of some words is limited to one or more dialects
was already taken into account in previous studies. For example it is known
that the word {\it alika} which refers to {\it dog} in Merina 
(the official variant) is replaced by the word {\it amboa} of African 
origin in most dialects.
Nevertheless, the study of Malagasy dialects in comparison with 
Indonesian languages is a still largely unexplored field of research.
Each dialect may provide pieces of information about the history the 
language,  eventually allowing us to for track the various linguistic 
influences experienced by Malagasy since the initial colonization of 
the Island.

An other open problem concerns the pre-Indonesian ancestral population.
It is still debated whether the island was inhabited before
the Indonesian colonization. In case the answer is positive it may be 
possible to track the aboriginal vocabulary in
the dialects. For example, the Mikea are the only hunter-gatherers 
in Madagascar, and it is unclear whether they are a relic of the
aboriginal pre-Indonesian population or just 'ordinary' Malagasy who
switched to a simpler economy for historical reasons.
If the first hypothesis is the correct one,
they should show some residual aboriginal vocabulary in their dialect, 
and the same is expected for the neighboring populations of 
Vezo and Masikoro.

\section*{Acknowledgments}

The ideas and the methods presented have been longly discussed
with Filippo Petroni, while important suggestions came from Philippe Blanchard,
Dimitri Volchenkov and S\"oren Wichmann.
We warmly thank Daniel Gandolfo, Eric W. Holman, Armando G. M. Neves and
Michele Pasquini for helpful discussions.
We also thank Joselin\`a Soafara N\'er\'e for her invaluable help in the 
collection of the Malagasy word lists.

\section*{Appendix A}

We start by our definition of lexical distance 
between two words, which is a variant of the Levenshtein distance \cite{Lev}. 
The Levenshtein distance is simply the minimum
number of insertions, deletions, or substitutions of a
single character needed to transform one word into the other.
Our distance is obtained by a normalization.

More precisely, given two words $\omega_1$ and $\omega_2$,
their distance $d(\omega_1, \omega_2)$ is defined as

\begin{equation}
d(\omega_1, \omega_2)= 
\frac{d_L(\omega_1, \omega_2)}{l(\omega_1, \omega_2)}
\label{wd}
\end{equation}
where $d_L(\omega_1, \omega_2)$ is their standard  Levenshtein distance
and $l(\omega_1, \omega_2)$ is the
number of characters of the longer of the two words
$\omega_1$ and $\omega_2$.
Therefore, the distance can take any value between 0 and 1.

The reason of the normalization can be understood by
the following example.
Consider the case of two words with the same length
in which a single substitution 
transforms one word into the other.  
If they are short, let's say 2 characters, they are very different. 
On the contrary, if they are long, let's say 8 characters, 
it is reasonable to say that they are very similar. 
Without normalization, their distance would be the same, equal to 1, 
regardless of their length. 
Instead, introducing the normalization factor, in the first 
case the distance is $\frac{1}{2}$,
whereas in the second, it is much smaller and equal to $\frac{1}{8}$. 

We use distance between pairs of words, as defined above, 
to construct the lexical distances of languages. 
For any language we prepare a list of words associated to the same 
$M$ meanings (we adopt the original Swadesh choice of $M=200$). 

Assume that the number of languages is $N$ and 
any language in the group is labeled by a Greek letter
(say $\alpha$) and any word of that language by 
$\alpha_i$ with $1 \leq i \leq M$. The same index $i$ 
corresponds to the same meaning in all languages i.e.,
two words $\alpha_i$ and $\beta_j$  
in the languages $\alpha$ and $\beta$ 
have the same meaning if $i=j$.

The lexical distance between two languages is then defined as 

\begin{equation}
D(\alpha, \beta)=  \frac{1}{M} \sum_{i=1}^M
d(\alpha_i, \beta_i)
\label{ld}
\end{equation}
It can be seen that $D(\alpha, \beta)$
is always in the interval [0,1] and obviously $D(\alpha, \alpha)=0$. 

The result of the analysis described above is a $N \times N$ upper 
triangular matrix whose entries are the $N(N-1)/2$ non-trivial 
lexical distances $D(\alpha, \beta)$ between all pairs of languages.

The matrix of the 23 Malagasy dialects, with entries multiplied by 1000, 
is the following:
\bigskip
\bigskip
 
\small
\setlength{\tabcolsep}{3.2pt}
\begin{tabular}{l | l*{22}{c}r}
 1 \\
 2 & 323 \\
 3 & 246 & 276 \\
 4 & 322 & 240 & 295 \\
 5 & 302 & 281 & 309 & 345 \\
 6 & 227 & 318 & 275 & 359 & 266 \\
 7 & 413 & 386 & 390 & 418 & 314 & 370 \\
 8 & 280 & 386 & 342 & 401 & 356 & 245 & 436 \\
 9 & 366 & 424 & 379 & 412 & 405 & 375 & 450 & 409 \\
10 & 411 & 396 & 416 & 440 & 318 & 366 & 249 & 456 & 482 \\
11 & 207 & 326 & 260 & 362 & 286 & 061 & 383 & 201 & 374 & 384 \\
12 & 362 & 343 & 345 & 387 & 292 & 328 & 289 & 397 & 435 & 330 & 324 \\
13 & 303 & 369 & 330 & 381 & 384 & 329 & 454 & 362 & 256 & 487 & 318 & 407 \\
14 & 343 & 302 & 331 & 355 & 243 & 317 & 303 & 403 & 423 & 314 & 336 & 301 & 419 \\
15 & 397 & 453 & 394 & 462 & 392 & 375 & 342 & 463 & 485 & 304 & 383 & 405 & 471 & 388 \\
16 & 368 & 391 & 385 & 416 & 392 & 390 & 448 & 406 & 320 & 474 & 383 & 429 & 325 & 418 & 486 \\
17 & 400 & 350 & 369 & 390 & 280 & 358 & 165 & 433 & 427 & 278 & 373 & 240 & 439 & 261 & 358 & 410 \\
18 & 322 & 376 & 325 & 374 & 391 & 337 & 426 & 381 & 198 & 473 & 339 & 412 & 234 & 406 & 461 & 264 & 414 \\
19 & 358 & 407 & 376 & 417 & 408 & 394 & 440 & 419 & 292 & 481 & 387 & 431 & 325 & 422 & 472 & 161 & 408 & 243 \\
20 & 297 & 388 & 359 & 430 & 356 & 299 & 400 & 346 & 386 & 433 & 275 & 375 & 363 & 375 & 455 & 348 & 394 & 349 & 355 \\
21 & 386 & 341 & 370 & 385 & 290 & 344 & 262 & 403 & 422 & 321 & 348 & 250 & 404 & 306 & 403 & 401 & 213 & 416 & 417 & 383 \\
22 & 225 & 389 & 332 & 394 & 382 & 316 & 471 & 319 & 385 & 475 & 287 & 421 & 296 & 431 & 480 & 382 & 467 & 348 & 387 & 356 & 441 \\
23 & 379 & 424 & 407 & 424 & 398 & 380 & 443 & 433 & 315 & 466 & 380 & 412 & 351 & 420 & 472 & 203 & 395 & 288 & 202 & 351 & 409 & 406 \\
\hline
 & 1 & 2 & 3 & 4 & 5 & 6 & 7 & 8 & 9 & 10 & 11 & 12 & 13 & 14 & 15 & 16 & 17 & 18 & 19 & 20 & 21 & 22 \\

\end{tabular}

\normalsize
\bigskip
\bigskip
\noindent
where the number-variant correspondence is:

\noindent
1 Antambohoaka (Mananjary), 
2 Antaisaka (Vangaindrano), 
3 Antaimoro (Manakara), 
4 Zafisoro (Farafangana), 
5 Bara (Betroka), 
6 Betsileo (Fianarantsoa), 
7 Vezo (Toliara), 
8 Sihanaka (Ambatondranzaka), 
9 Tsimihety (Mandritsara), 
10 Mahafaly (Ampanihy), 
11 Merina (Antananarivo), 
12 Sakalava (Morondava), 
13 Betsimisaraka (Fenoarivo-Est), 
14 Antanosy (Tolagnaro), 
15 Antandroy (Ambovombe), 
16 Antankarana (Vohemar), 
17 Masikoro (Miary), 
18 Antankarana (Antalaha), 
19 Sakalava (Ambanja), 
20 Sakalava (Majunga), 
21 Sakalava (Maintirano), 
22 Betsimisaraka (Mahanoro), 
23 Antankarana (Ambilobe).

\section*{Appendix B}

\noindent
Below we provide information on the people who furnished the data collected by the author at the beginning of 2010 with the invaluable help of 
Joselin\`a Soafara N\'{e}r\'{e}. For any dialect two consultants
have been independently interviewed.
Their names and birth dates follow each of the dialect names.

\begin{table}[p]
\caption{People which
 furnished the data on Malagasy dialects \label{Tab:1}}
\centering
\begin{tabular}[ht]{l l l}
\hline
\small \bf MERINA &\small  & \small \\
\small \bf(ANTANANARIVO)       
             & \small SERVA Maurizio &\small   \\ \hline
\small\bf ANTANOSY
             &  \small SOAFARA Joselina Nere & 
                                         \small  08 November 1987 \\
\small\bf  (TOLAGNARO)   &\small ETONO Imasinoro Lucia &\small  18 February 1982\\  \hline
\small\bf BETSIMISARAKA &
\small ANDREA Chanchette G\'{e}n\'{e}viane &\small  07 August 1985 \\
\small \bf  (FENOARIVO-EST)  &\small RAZAKAMAHEFA Joachim Julien   & \small 09 November 1977\\
\hline
\small \bf SAKALAVA &
 \small SEBASTIEN Doret & \small 26 November 1980 \\
 \small \bf  (MORONDAVA)  &\small RATSIMANAVAKY Christelle J.    & \small 29 February 1984\\
\hline
\small \bf VEZO &
\small RAKOTONDRABE Justin &\small  02 August 1972 \\
 \small \bf  (TOLIARA)  &RASOAVAVATIANA Claudia S.    &  \small 28 June 1983\\
\hline
\small\bf ZAFISIRO &
\small RALAMBO Alison & \small 11 June 1982 \\
\small \bf  (FARAFANGANA)  & \small RAZANAMALALA Jeanine    &\small  03 February 1980\\
\hline
\small \bf ANTAIMORO &
\small RAZAFENDRALAMBO Haingotiana & \small 24 July 1985 \\
\small \bf  (MANAKARA)  &\small RANDRIAMITSANGANA Blaise    & \small  05 February 1989\\
\hline
\small \bf ANTAISAKA &
\small RAMAHATOKITSARA Fidel Justin &\small  24 April 1984\\
\small \bf  (VANGAINDRANO)  & \small FARATIANA Marie Luise    &\small   17 August 1990\\
\hline
\small \bf ANTAMBOHOAKA &
\small RAKOTOMANANA Roger & \small 04 May 1979\\
\small  \bf  (MANANJARY)  &\small ZAFISOA Raly    & \small   20 April 1983\\
\hline
\small\bf BETSILEO &
\small RAMAMONJISOA Andrininina Leon Fidelis &\small 16 April 1987\\
 \small \bf  (FIANARANTSOA)  & \small RAKOTOZAFY Teza   &  \small 25 December 1985\\
\hline
\small\bf BARA &\small
RANDRIANTENAINA Hery Oskar Jean &\small 17 Jenuary 1986\\
\small \bf  (BETROKA)  & \small NATHANOEL Fife Luther  & \small  26 May 1983\\
\hline
\small\bf TSIMIHETY &
\small RAEZAKA Francis & \small 23 December 1984\\
\small \bf  (MANDRITSARA)  &\small FRANCINE Germaine Sylvia  &  \small 04 May 1985\\
\hline
\small \bf MAHAFALY &
\small VELONJARA Larissa &\small 21 April 1989\\
 \small \bf  (AMPANIHY)  &\small NOMENDRAZAKA Christian  & \small  07 June 1982\\
\hline
\small \bf SIHANAKA &
\small ARINAIVO Robert Andry & \small 06 Jenuary 1979\\
 \small \bf  (AMBATONDRAZAKA)  & \small RONDRONIAINA Natacha  &  \small 27 December 1985\\
\hline
\small\bf ANTANKARANA &
\small ANDRIANANTENAINA N. Benoit& \small 06 August 1984\\
\small \bf  (VOHEMAR)  &\small EDVINA Paulette & \small  28 Jenuary 1982\\
\hline
\small\bf ANTANKARANA &
\small RANDRIANARIVELO Jean Ives& \small 24 December 1986\\
\small \bf  (ANTALAHA)  &\small RAZANAMIHARY Saia &
 \small   07 September 1985\\
\hline
\small\bf SAKALAVA &
\small CASIMIR Jaozara Pacific& \small 03 April 1983\\
\small \bf  (AMBANJA)  &\small ZAKAVOLA M. Sandra &
 \small   17 July 1984\\
\hline
\small\bf SAKALAVA &
\small RATSIMBAZAFY Serge& \small 17 May 1978\\
\small \bf  (MAJUNGA)  &\small VAVINIRINA Fideline &
 \small    23 June 1970\\
\hline
\small\bf ANTANDROY &
\small RASAMIMANANA Z. Epaminodas& \small 05 June 1983\\
\small \bf  (AMBOVOMBE)  &\small MALALATAHINA Tiaray Samiarivola &
 \small    07 July 1984\\
\hline
\small\bf MASIKORO &
\small MAHATSANGA Fitahia& \small 22 March 1976\\
\small \bf  (ANTALAHA)  &\small VOANGHY Sidonie Antoinnette &
 \small    12 October 1981\\
\hline
\small\bf ANTANKARANA &
\small BAOHITA Maianne& \small 21 August 1984\\
\small \bf  (AMBILOBE)  &\small NOMENJANA HARY Jean Pierre Felix &
 \small    07 June 1980\\
\hline
\small\bf SAKALAVA &
\small HANTASOA Marie Edvige& \small 02 November 1985\\
\small \bf  (MAINTIRANO)  &\small KOTOVAO Bernard &
 \small    06 October 1983\\
\hline
\small\bf BETSIMISARAKA &
\small RASOLONANDRASANA Voahirana & \small 24 September 1985\\
\small \bf  (MAHANORO)  &\small ANDRIANANDRASANA Maurice &
 \small    03 April 1979\\
\hline
\end{tabular}
\end{table}

\hfill\eject

\begin{figure}
\epsfysize=16.0truecm \epsfxsize=18.0truecm
\centerline{\epsffile{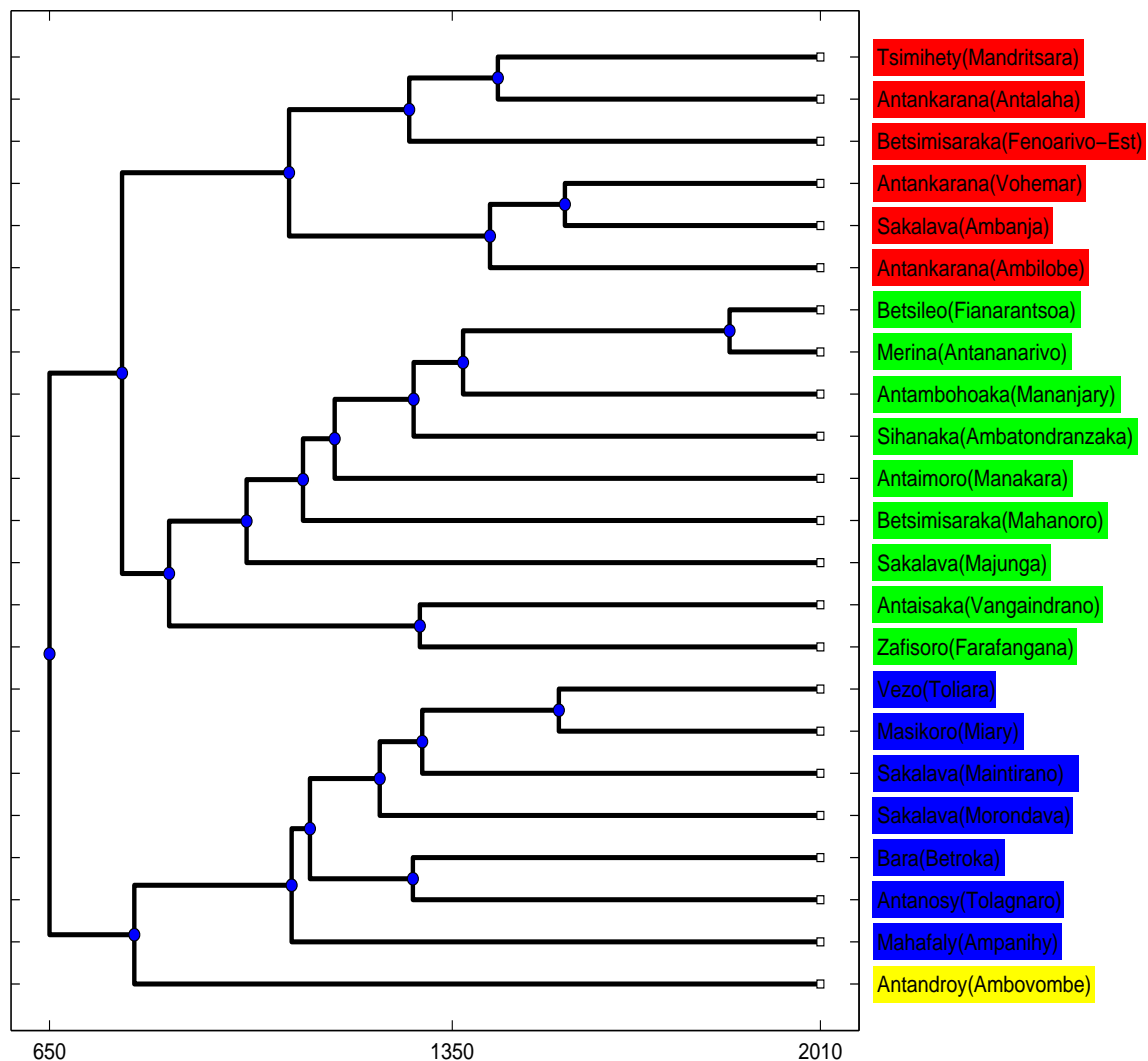}} 
\caption{Phylogenetic tree of 23 Malagasy dialects realized by
Unweighted Pair Group Method Average (UPGMA). 
In this figure the name of the dialect is followed
by the name of the town were it was collected.
The phylogenetic tree shows a main partition of the Malagasy dialects
into four main groups associated to different colors. 
The strict correspondence of this cladistic with the geography 
can be appreciate by comparison with Fig. 2.}
\label{fig1}
\end{figure}

\begin{figure}
\epsfysize=20.0truecm \epsfxsize=26.0truecm
\centerline{\epsffile{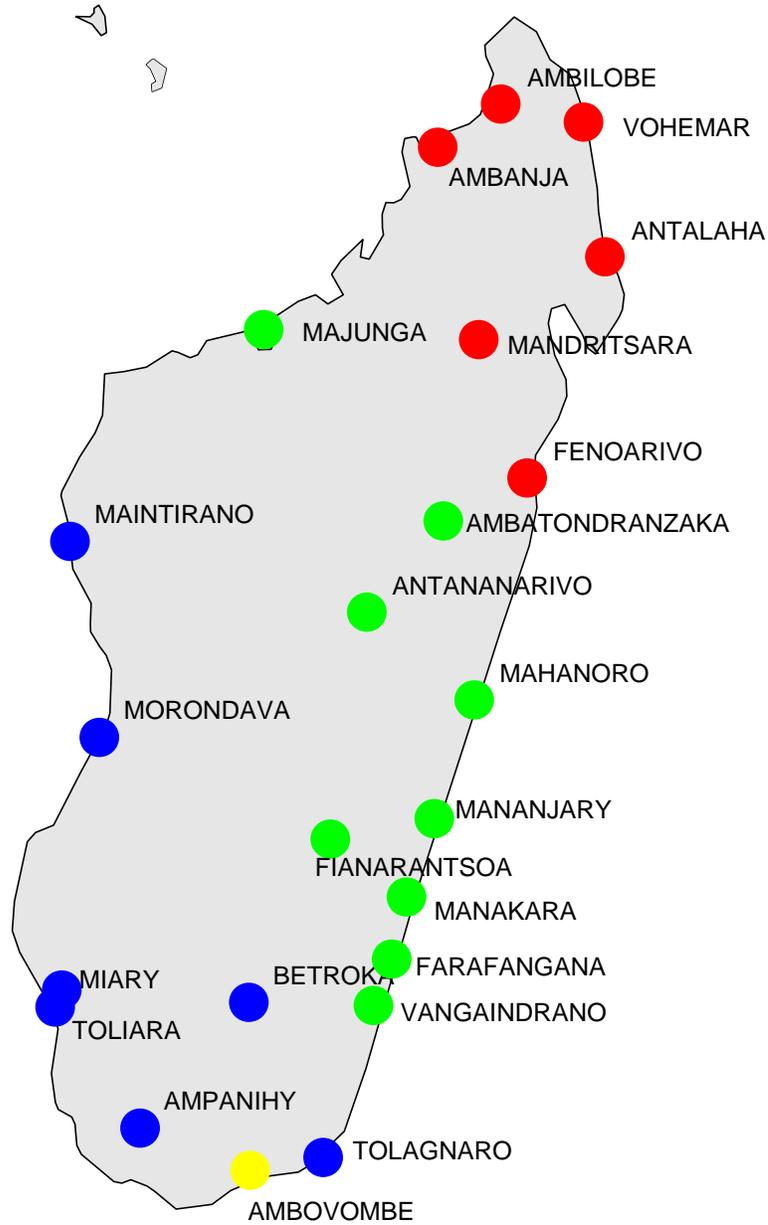}} 
\caption{Geography of Malagasy dialects.  The locations of the 23 dialects 
are indicated with the same colors of Fig. 1.
Any dialect is identified by the the name of the town where 
it was collected.}
\label{fig2}
\end{figure}

\begin{figure}
\epsfysize=16.0truecm \epsfxsize=18.0truecm
\centerline{\epsffile{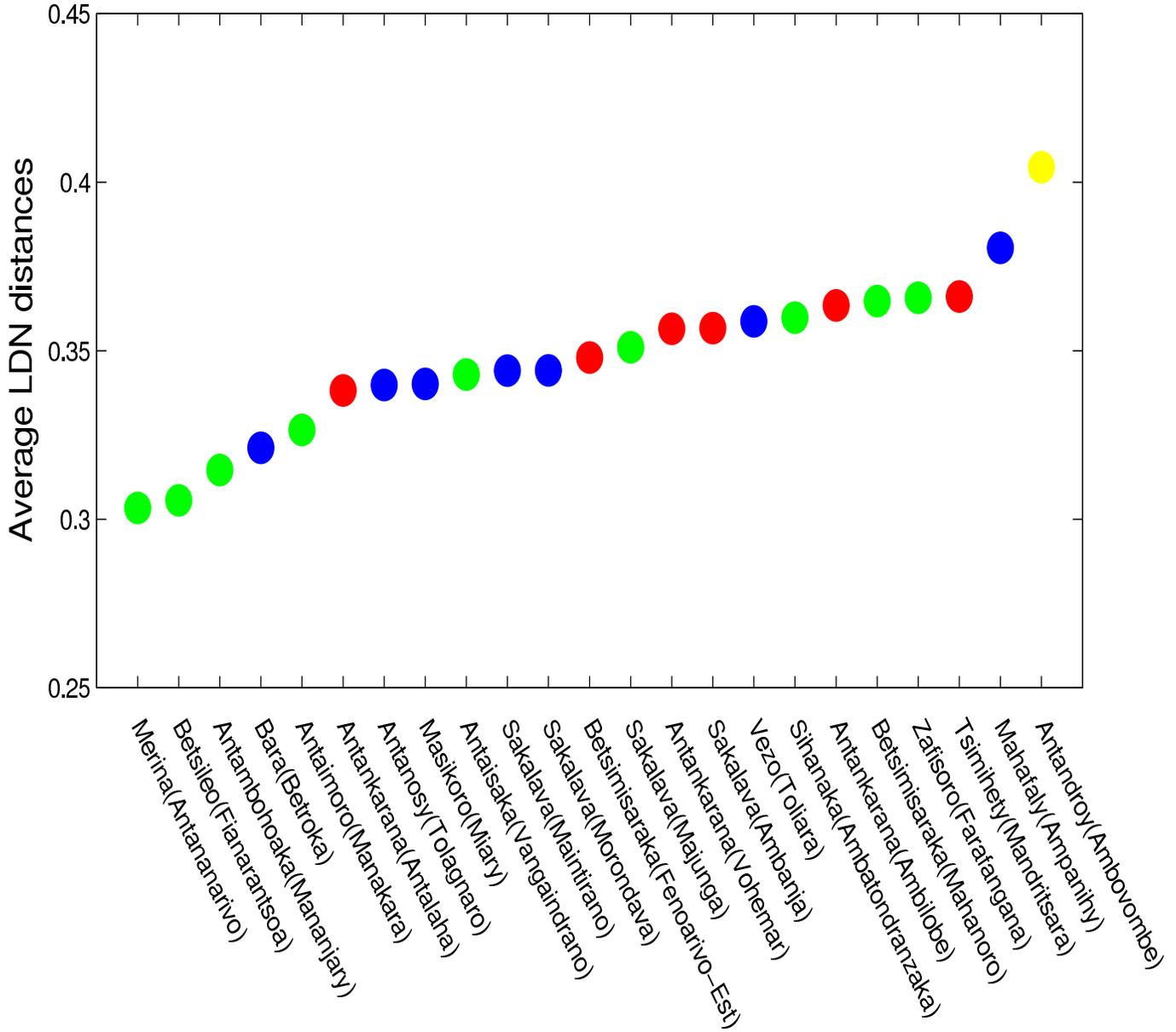}} 
\caption{Average distance of the Malagasy dialects from
all the others.  The 23 dialects are colored as in figure 2.
Highlands dialects (Antananarivo, Fianarantsoa and
Betroka) together with south-east coast dialects 
(Mananjary and Manakara) show the smallest average distance.}
\label{fig3}
\end{figure}

\begin{figure}
\epsfysize=20.0truecm \epsfxsize=20.0truecm
\centerline{\epsffile{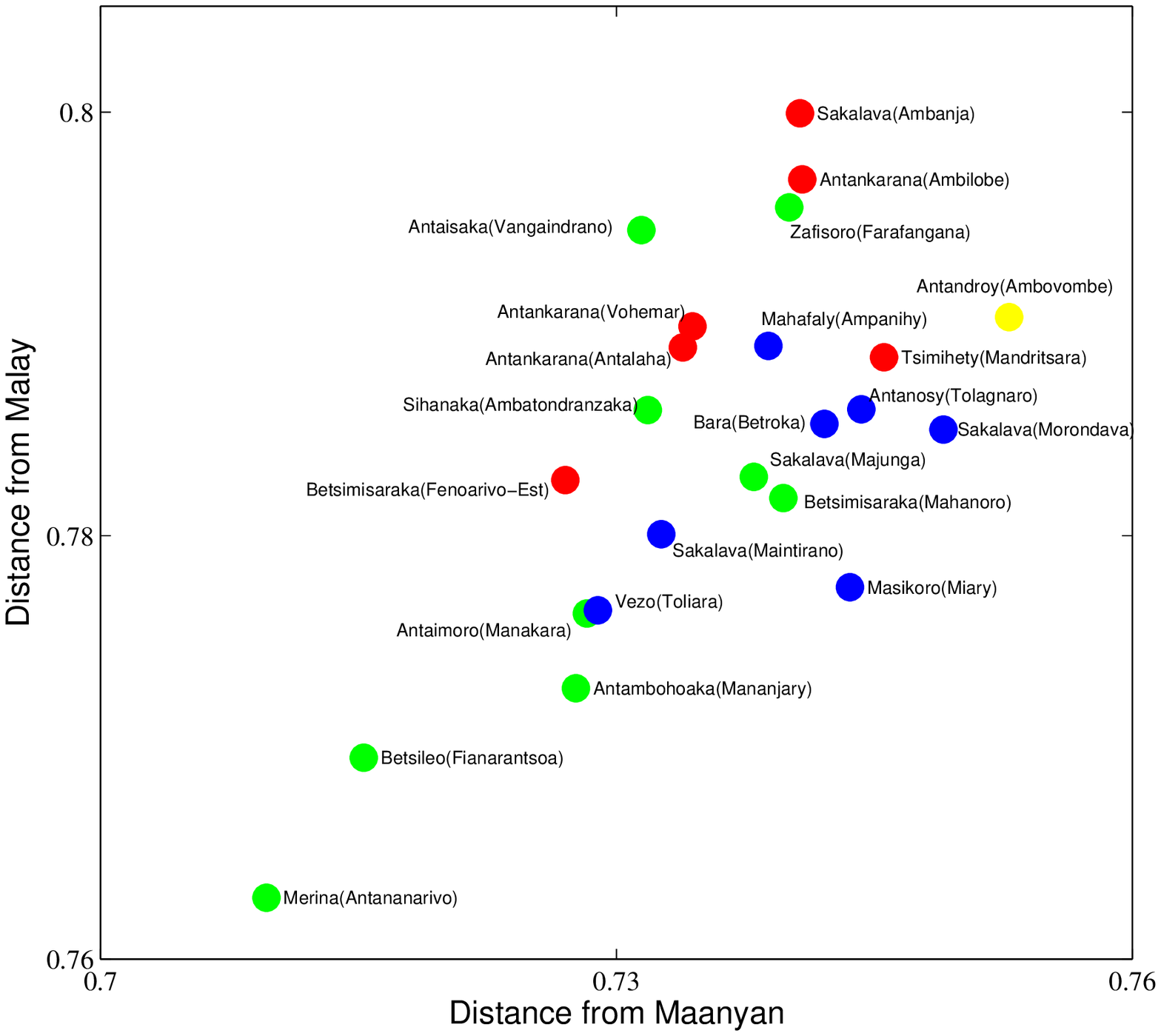}} 
\caption{Lexical distances of Malagasy dialects from Malay and Maanyan. 
The 23 dialects are indicated with the same colors of the map in Fig. 2.
Highlands dialects together with south-east coast Mananjary and Manakara 
dialects show the smallest distance from both the two Indonesian languages.}
\label{fig4}
\end{figure}

\end{document}